\documentclass[conference]{IEEEtran}
\usepackage{hyperref}

\usepackage{ifthen}
\newboolean{comments}
\setboolean{comments}{true} 
\ifthenelse{\boolean{comments}}{
\usepackage[textsize=tiny]{todonotes}
\newcommand{\zkn}[1]{\todo[color=orange!20, size=\tiny]{Zsolt: #1}}
\newcommand{\zk}[1]{{\textcolor{blue}{#1}}}
\setlength{\marginparwidth}{1.5cm}
}{
\newcommand{\zkn}[1]{{\textcolor{blue}{}}}
\newcommand{\zk}[1]{{\textcolor{blue}{}}}
}

\IEEEoverridecommandlockouts
\usepackage{cite}
\usepackage{amsmath,amssymb,amsfonts}
\usepackage{algorithmic}
\usepackage{graphicx}
\usepackage{afterpage}
\usepackage{textcomp}
\usepackage{xcolor}
\usepackage{booktabs}
\usepackage{multirow}
\usepackage{svg}
\usepackage{makecell}
\usepackage{caption}
\newboolean{anonymous}
\setboolean{anonymous}{false} %
\usepackage[letterpaper, top=1in, bottom=0.75in, left=0.75in, right=0.75in]{geometry}
\usepackage{float}
\usepackage{stfloats}

\def\BibTeX{{\rm B\kern-.05em{\sc i\kern-.025em b}\kern-.08em
    T\kern-.1667em\lower.7ex\hbox{E}\kern-.125emX}}

    \ifthenelse{\boolean{anonymous}}{%
  \let\thanks\relax
  \renewcommand{\and}{} %
  \usepackage{hyperref}
  \hypersetup{pdfauthor={},pdfcreator={},pdftitle={}}
  \author{\vspace{-0.5em}%
    \begin{minipage}{\linewidth}\centering
    \textbf{Anonymous Submission}\\[2pt]
    \textit{Paper under double-blind review}
    \end{minipage}\vspace{-0.5em}
  }
}{%
  \author{%
    Jeremiah Coholich\textsuperscript{1}, 
    Justin Wit\textsuperscript{1}, 
    Robert Azarcon\textsuperscript{1}, 
    Zsolt Kira\textsuperscript{1}%
    \thanks{\textsuperscript{1}Institute of Robotics and Intelligent Machine, Georgia Institute of Technology, Atlanta, GA, USA. Emails: \{jcoholich, jwit3, razarcon3, zkira\}@gatech.edu.}
  }}
\begin{document}

\title{\looseness-1 Sim2real Image Translation Enables Viewpoint-Robust Policies from Fixed-Camera Datasets}

\maketitle

\begin{abstract}
\looseness=-1 Vision-based policies for robot manipulation have achieved significant recent success, but are still brittle to distribution shifts such as camera viewpoint variations. Robot demonstration data is scarce and often lacks appropriate variation in camera viewpoints. Simulation offers a way to collect robot demonstrations at scale with comprehensive coverage of different viewpoints, but presents a visual sim2real challenge. To bridge this gap, we propose MANGO -- an unpaired image translation method with a novel segmentation-conditioned InfoNCE loss, a highly-regularized discriminator design, and a modified PatchNCE loss. We find that these elements are crucial for maintaining viewpoint consistency during sim2real translation. When training MANGO, we only require a small amount of fixed-camera data from the real world, but show that our method can generate diverse unseen viewpoints by translating simulated observations. In this setting, MANGO outperforms all other image translation methods we tested. In certain real-world tabletop manipulation tasks, MANGO augmentation increases shifted-view success rates by over 40 percentage points compared to policies trained without augmentation. For more results, visit: \url{https://www.jeremiahcoholich.com/mango}.

\end{abstract}

\section{Introduction}
\label{sec:intro}
Significant progress has been made in developing vision-based imitation-learning policies for robot manipulation. Performant single-task architectures ~\cite{act, wang2024scaling, ke20243d, chi2023diffusion} and intuitive teleoperation methods \cite{openteach, act, wu2024gello} have given way to large robot datasets and multi-task Vision-Language-Action models (VLAs) ~\cite{kim2024openvla, szot2025grounding, black2024pi_0, intelligence2025pi_, bjorck2025gr00t}. However, the robot demonstrations used to train these models are scarce and labor-intensive to generate. In comparison to web-scraped vision-language datasets, robot learning datasets often lack diversity which results in models with poor zero-shot performance to new setups.

\begin{figure}[t]
    \centering
    \includegraphics[width=1.0\linewidth]{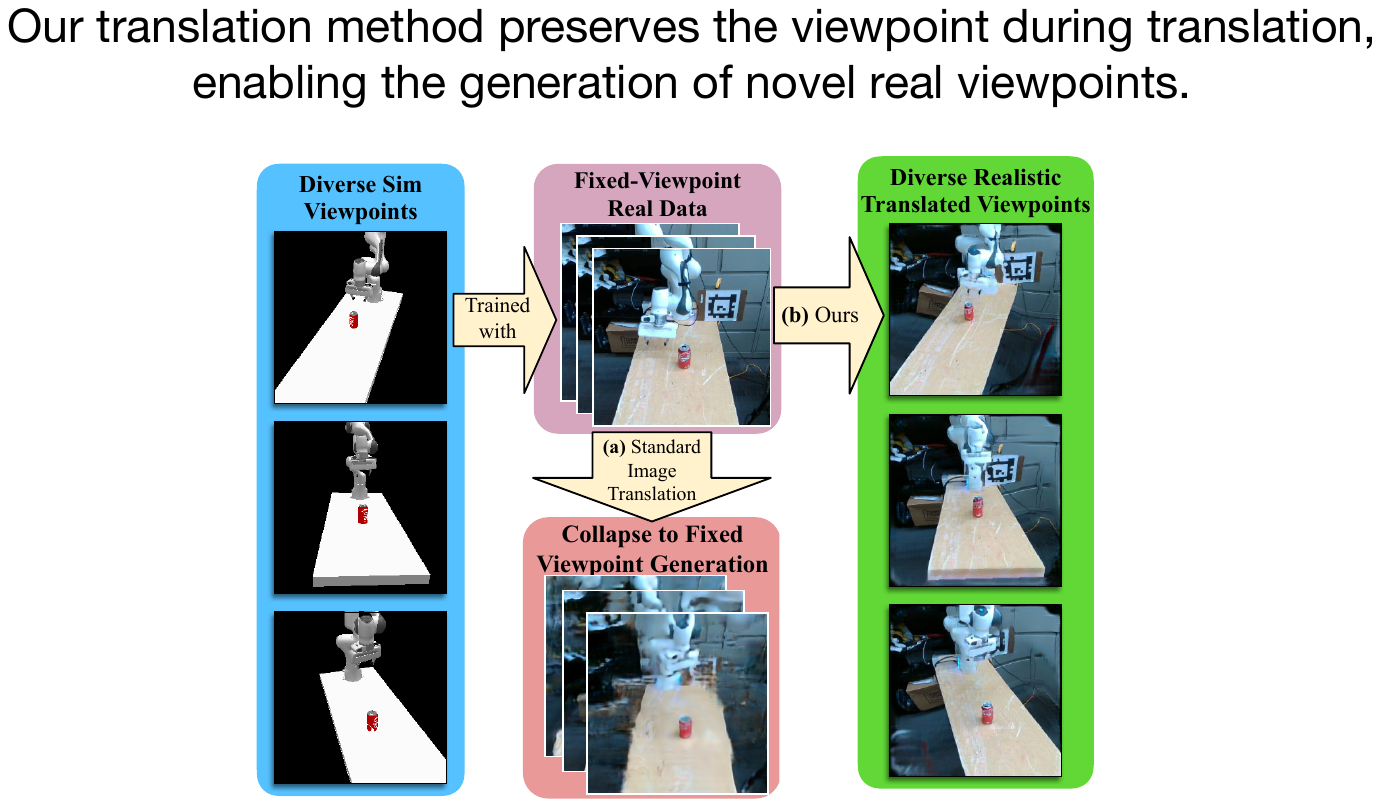}
    \caption{(a) Standard image translation methods fail to generalize to new viewpoints when trained on a fixed-viewpoint target domain dataset. (b) Our method, MANGO, enables realistic generation of unseen viewpoints, which are used to improve the robustness of downstream robot polices.}
    \label{fig:intro}
\end{figure}

For example, many tabletop manipulation datasets employ fixed, third person cameras for observations \cite{haldar2023watch, bridgedata, lee2019making, wilcox2025adapt3r, ke20243d, chi2023diffusion, ze20243d}, making downstream policies brittle to camera viewpoint shifts. Indeed, we have observed that when robot polices are trained on fixed-camera datasets, changes in camera viewpoint during deployment cause success rates to crater (Table \ref{tab:real_results}). Cameras are often fixed during demo collection to ensure consistent visual observations, avoid repeated calibration with depth or motion capture sensors, or simply for convenience. Generalizing to truly unseen viewpoints is difficult because a change in viewpoint affects the entire scene, and the model must implicitly estimate the robot's position relative to the new camera position.

To augment fixed-camera robot data, we propose to collect simulated demonstrations from a simple digital twin via task and motion planning whose visual observations are taken from diverse camera viewpoints. Crucially, we train a novel image translation model for bridging the visual sim2real gap. We name our proposed approach Multiview Augmentation with Novel Generated Observations, or MANGO. With MANGO, we simultaneously bypass manual data collection and solve the viewpoint diversity issue. Our MANGO image translation model is trained on a small real-world dataset collected from a single fixed camera plus a larger simulation dataset of images with segmentations. After training, MANGO is able to translate diverse simulation viewpoints to unseen real-world viewpoints. While our method relies on a digital twin, we use a deliberately simple simulation rendered with low-fidelity OpenGL settings, without extensive visual engineering. The entire pipeline yields synthetic demonstrations with realistic and varied camera viewpoints.
\begin{figure*}[ht!]
    \centering
    \includegraphics[width=1.0\linewidth]{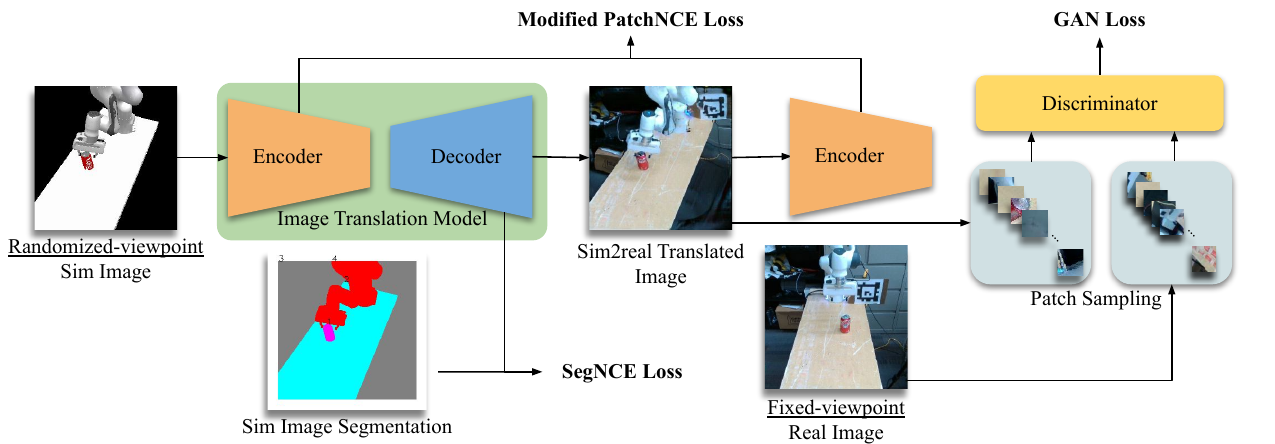}
    \caption{MANGO is trained on unpaired real and sim images, specifically a real dataset obtained with a fixed camera and a simulation dataset with diverse camera viewpoints. To ensure the simulation viewpoint is preserved during translation, we employ a novel segmentation-based InfoNCE loss, a modified PatchNCE loss from \cite{CUT}, and a random patch sampling and rotation process to regularize the discriminator $D$.}
    \label{fig:method}
\end{figure*}

We argue for GAN-based models over diffusion models for robot data augmentation. Robot demonstration datasets contain many image observations which are downsampled to small image sizes. A robot dataset for a small, single-task policy may contain upwards of 180,000 images, assuming 150 demonstrations collected at 20 Hz. Using ZeroNVS \cite{sargent2024zeronvs}, a state-of-the-art diffusion model for novel viewpoint generation, augmenting this dataset would take 435 GPU-hours. Training an ACT policy on such a dataset would only require $\sim 5$ hours on a single GPU, meaning that diffusion model-based augmentation would severely bottleneck compute requirements. Our lightweight GAN-based approach is approximately 2,700x times faster than ZeroNVS, making it practical for visual dataset augmentation.

\noindent Our core contributions are as follows:

\begin{enumerate}
    \item MANGO: A sim2real image translation method incorporating a novel, segmentation-informed contrastive InfoNCE loss capable of preserving unseen simulation viewpoints during translation
    \item Experimental proof that demonstrations generated with MANGO improves downstream robot policy robustness to shifts in camera position, even in comparison to a diffusion-based method
   \item Analysis of why our method is successful on the domain of robot demonstration datasets in comparison to the many other approaches developed for the generic problem of unpaired image translation
\end{enumerate}

\section{Related Work}

\subsection{Visual Sim2Real Translation for Robotics}

Simulation is a popular tool among roboticists, but to actually leverage simulated data for real robot policies requires bridging the sim2real gap. For visual observations, one option is to improve simulator realism \cite{li2024evaluating}, however this is a labor-intensive engineering effort which must be done for every scene. Domain-randomization of lighting, colors, and textures is another option \cite{pinto2017asymmetric, andrychowicz2020learning, garcia2023robust}, but determining the degree and types of randomizations to apply is a challenge, and policies trained with domain randomization often sacrifice performance for robustness. 

Image-to-image translation has been proposed to cross this sim2real gap by learning from data. A wide variety of unpaired image-to-image translation architectures exist \cite{cyclegan, CUT, zhao2022egsde, choi2018stargan, yi2017dualgan, sushko2020you, choi2021ilvr}. To translate from sim2real, roboticists can directly train these models on datasets of image observations collected from simulation and the real world. For example, in \cite{zhang2023reinforcement} and \cite{truong2021bi}, the authors train an unmodified CycleGAN to translate visual observations for grasping and navigation.

Others have tailored these methods to incorporate specific knowledge about the robot and downstream application. DigitalTwin-CycleGAN adds an action cycle-consistency loss to CycleGAN for a sim2real visual grasping task \cite{dt-cyclegan}. This loss makes the image translation model dependent on learning a successful grasping policy concurrently. RL-CycleGAN incorporates Q-function consistency on translated images\cite{rl-cyclegan}, where the Q-function is obtained while learning a task-specific RL policy. RetinaGAN enforces cycle consistency with an object detector which requires thousands of labeled images to train beforehand \cite{retinagan}. GraspGAN trains an image translation model without cycle-consistency and instead enforces accurate image content translation through a grasp success predictor \cite{graspgan}. Additionally, they include an auxiliary generator objective of reproducing the ground-truth sim image segmentation. CyCADA unifies these methods under a general “task loss” framework \cite{hoffman2018cycada}. In contrast, MANGO is agnostic to the downstream learning algorithm, enabling us to train one image translation model for many tasks. 

Diffusion models \cite{ho2020denoising} have emerged as the primary architecture for image generation over generative adversarial networks (GANs) \cite{gan}, with some exceptions \cite{yu2022scaling, kang2023scaling, sauer2023stylegan}. However, we find that for the specific domain of unpaired image-to-image translation, GANs obtain results competitive with the best diffusion approaches \cite{zhao2022egsde}. We hypothesize that that the output multimodality of diffusion models is a disadvantage when the style and content of the generated image are tightly-specified by the input image and target domain dataset, respectively. MANGO uses the GAN loss; however in theory our novel segmentation-based InfoNCE loss could be applied to any image translation architecture containing a generator network with spatially-indexed latent feature maps.

\subsection{Robot Policy Viewpoint Invariance}
RoboNet offered early proof that training a robot policy on multiple views enables generalization to viewpoint shifts \cite{robonet}. Multi-view Masked World Models (MV-MWM) \cite{seo2023multi} demonstrates impressive robustness to camera viewpoints by training a viewpoint-invariant visual encoder and task-specific world model. MoVie \cite{yang2023movie} achieves view generalization by adapting an image encoder to the novel views during test-time. In contrast, MANGO does not require any test-time adaptation or real-world images from viewpoints outside of the fixed-camera images used for training. \cite{acar2023visual} trains an RL policy that is robust to single-camera viewpoint changes after learning from a teacher policy trained with a multi-view observation. VISTA leverages pretrained models with 3D priors to generate novel viewpoints given a single real-world image observation \cite{tian2024view}. However, since they do not use simulation-generated demonstrations they are unable to generate new robot trajectories and must rely on human demonstration collection. Additionally, VISTA finetunes the ZeroNVS \cite{sargent2024zeronvs} diffusion model, which suffers from high resource requirements as discussed in Section \ref{sec:intro}.

Learned 3D representations are inherently viewpoint invariant in theory, but still overfit to the specific 2D sensor locations in practice. Additionally, building strong 3D implicit or explicit representations typically requires more data than a single RGB sensor. For example, GROOT \cite{groot} achieves impressive viewpoint invariance but requires task-specific object annotations. 3D Diffusion Policy and 3D Diffusor-Actor both build 3D scene representations from calibrated RGBD cameras, but are shown to be brittle to the viewpoints used for this synthesis \cite{ze20243d,ke20243d}. Adapt3R achieves greater viewpoint robustness through only mapping embedding vectors to 3D instead creating entire scene pointclouds, but still requires multiple calibrated RGBD external sensors\cite{wilcox2025adapt3r}. MANGO demonstrations viewpoint robustness with only a single external RGB camera.

\section{Method}
\subsection{Image-to-image Translation}
We propose MANGO, a novel unpaired image-to-image translation method to translate visual observations from sim to real (Figure \ref{fig:method}). MANGO is trained on a small, fixed-viewpoint dataset of real images yet is capable of accurately translating viewpoint-diverse observations from a simple digital twin to realistic unseen viewpoints.

The objective of unpaired image-to-image translation is to translate images from domain $A$ to domain $B$ without access to a paired dataset of images $\mathcal{D}_{paired} = \{d_A,\; d_B |d_A \in A,\; d_b \in B \}_{i=0}^N$. Instead, we learn from two disjoint datasets $\mathcal{D}_A$ and $\mathcal{D}_B$, where our domain $A$ is simulation, our domain $B$ is the real-world, and $|\mathcal{D}_A| > |\mathcal{D}_B|$. This problem is considered unsupervised because there is no label, or ground-truth image, in $\mathcal{D}_B$ that images in $\mathcal{D}_A$ map to.

Image translators like MANGO must change the style of the input image while preserving its content. We employ the GAN architecture with a highly-regularized discriminator to learn the target domain style. For accurate content preservation, we use the InfoNCE \cite{infonce} loss between input and output image features in a similar style as CUT \cite{CUT}, but with a modified scoring function. Additionally, we propose a novel segmentation-based InfoNCE loss on generator features.

\subsection{Style Loss}

We use the standard GAN loss \cite{gan} to enforce target domain style on the generated images, given by Equation \ref{eq:gan}. $G$ is the generator network, and $D$ is the discriminator network.
\begin{equation}
\begin{split}
\label{eq:gan}
    \mathcal{L}_{\text{GAN}}(&G, D, \mathcal{D}_A, \mathcal{D}_B) = \mathbb{E}_{x \sim \mathcal{D}_B} \log(D(x)) \\&+\ \mathbb{E}_{y \sim \mathcal{D}_A} \log(1 - D(G(y)))
\end{split}
\end{equation}
One assumption underlying image-translation GANs, such as CycleGAN\cite{cyclegan} and CUT \cite{CUT}, is that the shared attributes among all images in $\mathcal{D}_B$ constitute the target domain ``style''. However, our real-world robot image observations in $\mathcal{D}_B$ only differ from one another in robot and object poses. Much of the image content, such as the background and tabletop, is nearly identical in all images in $\mathcal{D}_B$. A na\"ive discriminator will memorize the repetitive details and force the generator to recreate them. To mitigate this problem, Pix2pix \cite{pix2pix} proposed a ``PatchGAN'', where the discriminator only receives local image patches and cannot therefore enforce global image elements. We take this a step further and randomly sample patch locations and apply per-patch random rotations. This process is shown in Figure \ref{fig:method}. The result is a highly-regularized discriminator capable of enforcing the style of $\mathcal{D}_B$ on images with viewpoints not seen in $\mathcal{D}_B$.

\begin{figure}
    \centering
    \includegraphics[width=0.75\linewidth]{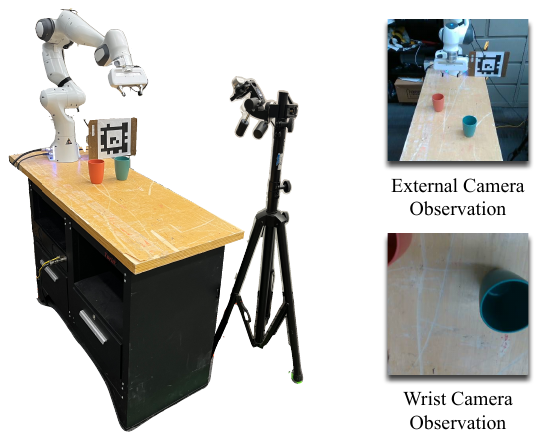}
    \caption{Our real-world robot setup, including a Franka Emika Panda arm, a wrist camera, and an external camera that is only repositioned for evaluations.}
    \label{fig:setup}
\end{figure}
\subsection{Content Loss}
MANGO content translation losses consists of a modified version of the PatchNCE loss \cite{CUT} and a novel segmentation-based NCE loss.

\subsubsection{Modified PatchNCE Loss}

The PatchNCE loss consists of an InfoNCE loss across encoder features generated by a source domain image $d_A \in D_A$ and its corresponding translated output image $G(d_A)$. For an input image $d$, we randomly sample $N$ latent features from the encoder at $L$ different layers. We call the set of features at layer $l$ $\mathcal{Z}_l$ and $|\mathcal{Z}_l| = N \ \forall\  l \in \{l_0,...,l_L\}$. The translated image $\hat{d}$ is passed through the encoder again to obtain $\hat{\mathcal{Z}}_l\ \forall\  l \in \{l_0,...,l_L\}$. All $\hat{\mathcal{Z}}_l$ are obtained from the same feature map indices as in  $\mathcal{Z}_l$.

The InfoNCE loss for feature $i$ in encoder layer $l$ is given by Equation \ref{eq:patchNCE}. This is the categorical cross-entropy loss on the probability that a feature $\mathbf{z} \in \mathcal{Z}$ will be correctly classified as the corresponding feature in $\hat{\mathcal{Z}}$, based on a scoring function $\rho_l(\cdot)$. $\tau$ is a temperature hyperparameter.
\vspace{-5pt}
\begin{equation}
    \ell_{\text{NCE}}(l, \mathbf{z}, \hat{\mathcal{Z}}, i) = -\log\left[\frac{\exp\left(\rho_l(\mathbf{z}, \hat{\mathbf{z}}_i) / \tau\right) }{ \displaystyle{\sum\limits_{\hat{\mathbf{z}} \in \hat{\mathcal{Z}}} \exp\left(\rho_l(\mathbf{z}, \hat{\mathbf{z}})/ \tau\right)}}\right]
    \label{eq:patchNCE}
\end{equation}

 $\rho_l(\cdot)$ is defined in Equation \ref{eq:scoring}. Features $\mathbf{z}_i$ and $\mathbf{z}_j$ are passed through a function $H_l$ then scored with cosine similarity.
\vspace{-2pt}
\begin{equation}
\label{eq:scoring}
\rho_l(\mathbf{z}_i, \mathbf{z}_j) = \frac{H_l(\mathbf{z}_i) \cdot H_l(\mathbf{z}_j)}{\|H_l(\mathbf{z}_i)\| \| H_l(\mathbf{z}_j)\|}
\end{equation}

The full loss is given in Equation \ref{eq:full_patchNCE}
\vspace{-3pt}
\begin{equation}
\label{eq:full_patchNCE}
    \mathcal{L}_{\text{PatchNCE}}(G, H, \mathcal{D}) = \mathbb{E}_{d \sim \mathcal{D}}\sum_{l=1}^L \sum_{i=1}^{|\mathcal{Z}_l|} \ell_{\text{NCE}}(l,\ \mathbf{z}_{l,i},\ \hat{\mathcal{Z}}_l,\ i)
\end{equation}

The reasoning behind Equation \ref{eq:patchNCE} is that input and output features from the same feature map locations are ``positive'' samples and should have high similarity scores. All other features are ``negative'' samples and should be repelled. However, we observe that many different input image patches are highly similar due to repeated patterns or textures in robot image datasets, which include background elements, the tabletop, etc. Furthermore, in the simulated image dataset $\mathcal{D}_A$, these regions all have \textit{identical} pixel values due to simplistic rendering. Therefore, Equation \ref{eq:patchNCE} will repel many false negative features.

To mitigate this, we train MANGO with a modified scoring function. If the cosine similarity of a negative sample exceeds a threshold $\theta$, we multiply the score by a factor $0 \leq \alpha < 1$. The modified scoring function is given by Equation \ref{eq:modified_scoring}.

\begin{equation}
\label{eq:modified_scoring}
\tilde{\rho}_l(\mathbf{z}_i, \mathbf{z}_j) =
\begin{cases}
\alpha \, \rho_l(\mathbf{z}_i, \mathbf{z}_j) & \text{if } \rho_l(\mathbf{z}_i, \mathbf{z}_j) > \theta\ \text{and}\ i \neq j, \\
\rho_l(\mathbf{z}_i, \mathbf{z}_j) & \text{otherwise.}
\end{cases}
\end{equation}

We have empirically found this to be more effective than increasing $\tau$. We denote the modified NCE loss which uses the scoring function in Equation \ref{eq:modified_scoring} as $\tilde{\mathcal{L}}_{\text{PatchNCE}}$. In practice, we set $\alpha=0.5$ and $\theta=0.9$.

\subsubsection{Segmentation NCE Loss}

We leverage the simulator used to generate $\mathcal{D}_A$ to obtain ground-truth segmentation maps for each image. We propose an InfoNCE loss which clusters generator features by segmentation category in order to ensure that object boundaries are preserved during translation. We note that for object-centric manipulation, this is a crucial aspect to preserve.

\looseness=-1 Each image in $\mathcal{D}_A$ contains $C$ segmentation classes and each feature $\mathbf{z}_i \in \mathcal{Z}$ generated from $d \sim D_A$ has an associated class label $y_i \in \mathcal{Y}$. In the case when a layer's feature map is of a lower resolution than the input image, we scale the image segmentation with nearest-neighbors downsampling to obtain $y_i$. 

The Segmentation NCE (SegNCE) loss is defined in Equation \ref{eq:segNCE_small}. In contrast to $\ell_{\text{NCE}}$ as shown in Equation \ref{eq:patchNCE}, there are multiple positive samples for the query feature $\mathbf{z}_i$. All features that are in the same segmentation class as the query feature are classified as positive samples and indexed by $j$.

In Equation \ref{eq:patchNCE}, the target distribution for the cross-entropy loss is a one-hot vector. In Equation \ref{eq:segNCE_small} the target distribution is a uniform distribution over features from the same segmentation class and zero elsewhere.

Here, we use the original scoring function $\rho_l(\cdot)$ defined in Equation \ref{eq:scoring}; since we are operating with ground-truth image segmentations, there are no false negatives.

\vspace{-3pt}
\begin{equation}
    \label{eq:segNCE_small}
    \begin{split}
    \ell_{\text{SegNCE}}(l, \mathcal{Z}, i, \mathcal{Y}) = \ \ \ \ \ \ \ \ \ \ \ \ \ \ \ \ \ \ \ \ \ \ \ \ \ \ \ \ \ \\
    \frac{1}{\left|\left\{ j \middle|\substack{j\in 1..|\mathcal{Z}|\\ y_j = y_i\\ i \neq j} \right\}\right|} \sum_{\left\{ j \middle|\substack{j\in 1..|\mathcal{Z}|\\ y_j = y_i\\ i \neq j} \right\}}  \ell_{\text{NCE}}(l, \mathbf{z}_i, \mathcal{Z}, j)
    \end{split}
\end{equation}

The full loss term is given in Equation \ref{eq:full_segnce}.

\vspace{-5pt}
\begin{equation}
\label{eq:full_segnce}
    \mathcal{L}_{\text{SegNCE}}(G, H, \mathcal{D}_A) = \mathbb{E}_{d\sim \mathcal{D}_A}\sum_{l=1}^L \sum_{i=1}^S \ell_{\text{SegNCE}}(l, \mathcal{Z}_l, i, \mathcal{Y}_l)
\end{equation}

The SegNCE loss is computed from input image generator features only.

  \subsection{Model training }

  The total loss function for $G$ is given in equation \ref{eq:total_G_loss}. We include an identity PatchNCE loss for regularization following \cite{CUT}. The full discriminator loss is given in Equation \ref{eq:total_D_loss} and is simply the GAN objective.

  \begin{equation}
  \label{eq:total_G_loss}
  \begin{split}
      \mathcal{L}_G = \tilde{\mathcal{L}}_{\text{PatchNCE}}(G, H, \mathcal{D}_A) \\
      + \tilde{\mathcal{L}}_{\text{PatchNCE}}(G, H, \mathcal{D}_B) \\
      + \mathcal{L}_{\text{SegNCE}}(G, H, \mathcal{D}_A) \\
      +\mathcal{L}_{\text{GAN}}(G, D, \mathcal{D}_A, \mathcal{D}_B)
  \end{split}
  \end{equation}

  \begin{equation}
    \label{eq:total_D_loss}
      \mathcal{L}_D = -\mathcal{L}_{\text{GAN}}(G, D, \mathcal{D}_A, \mathcal{D}_B)
  \end{equation}

\section{Experiments}

Our experiments are designed to answer the following questions:
\begin{enumerate}
    \item How well can MANGO translate sim images to unseen unseen real-world viewpoints? 
    \item Are imitation-learning policies trained with synthetic data from MANGO more robust to shifts in camera position?
    \item How does MANGO compare to baselines such as domain randomization and diffusion-based image augmentation methods? 
\end{enumerate}

\subsection{Image Translation with MANGO}

\subsubsection{Training Details}

Our generator $G$ is a 12M parameter ResNet-based network. The discriminator $D$ is a wider three-layer CNN with 11M parameters. Additionally, we parameterize the $H_l$ in Equations \ref{eq:scoring} and \ref{eq:modified_scoring} as a two-layer MLP with 700k parameters.

We first benchmark the image translation method on observations from a ``\texttt{pick up coke}'' task. $\mathcal{D}_A$ is a dataset of 8,098 image observations from simulation with camera viewpoints randomized within a box of dimensions (100, 100, 84) cm (L$\times$W$\times$H). $\mathcal{D}_B$ contains 3,094 images obtained from roughly 10 minutes of real-world teleoperated play data. All images are cropped and scaled to 256x256. Training the image translation model takes approximately 20 hours on a single RTX 2080 Ti GPU.

We curate three test datasets: Fixed View, Randomized View, and Wrist View. Each test set contains 128 sim/real image pairs. The Fixed View test set contains sim and real images from the same fixed view as $\mathcal{D}_B$. The randomized view testset contains both sim and real viewpoints taken from a camera placed randomly within a box of dimensions (100, 100, 84) cm. In order to create paired images for the Randomized View testset, we use the robot state in conjunction with AprilTags for coke can and camera pose estimation. Note that these are not needed for the deployment of MANGO, only for test set creation.

\begin{table}[h]
\centering
\caption{\textbf{Sim2real Unpaired Image Translation FID Scores.} Each score is averaged from two models trained with different seeds. $D$ is the discriminator and $\mathcal{D}_A$ is the sim training image dataset.}
\label{tab:benchmark_results}
\begin{tabular}{lccc}
\toprule
\textbf{Method} & \makecell{\textbf{Fixed View}\\\textbf{FID}($\downarrow$)} & \makecell{\textbf{Randomized }\\\textbf{View FID}($\downarrow$)} & \makecell{\textbf{Wrist View}\\\textbf{FID}($\downarrow$)} \\
\midrule
No Translation & 340.3 & 297.4 & 268.8 \\
\midrule
CUT \cite{CUT} & 412.3 & 373.9 & 266.3 \\
CycleGAN \cite{cyclegan} & 393.9 & 359.9 & 265.8 \\
\midrule
Basic $D$ & 371.3 & 318.5 & 293.5 \\
Without SegNCE loss & 267.0 & 207.5 & 199.9 \\
Without $\tilde{\rho}_l(\cdot)$ & 192.9 & 184.1 & 195.3 \\
Fixed-cam $\mathcal{D}_A$ & \textbf{108.0} & 198.7 & 202.5 \\
MANGO & 182.3 & \textbf{160.9} & \textbf{191.3} \\
\bottomrule
\end{tabular}
\end{table}

\subsubsection{Results}
\begin{table*}[h!]
\centering
\caption{\textbf{Sim2sim Experiment Results.} Success rates are the average of 50 rollouts. FID scores are computed across target-domain generated images and target-domain oracle images.}
\label{tab:sim_results}
\begin{tabular*}{\textwidth}{@{\extracolsep{\fill}}l*{6}{c@{}c}}
\toprule
 & \multicolumn{4}{c}{\textbf{Unseen Object}} 
 & \multicolumn{4}{c}{\textbf{Shared Object}} 
 & \multicolumn{4}{c}{\textbf{Cross-Embodiment}} \\
\cmidrule(lr){2-5} \cmidrule(lr){6-9} \cmidrule(lr){10-13}
\textbf{Data Augmentation}
 & \textbf{Threading$\uparrow$} & \textit{FID$\downarrow$}
 & \textbf{Hammer$\uparrow$} & \textit{FID$\downarrow$}
 & \textbf{Coffee$\uparrow$} & \textit{FID$\downarrow$}
 & \textbf{Stack$\uparrow$} & \textit{FID$\downarrow$}
 & \textbf{PickPlace$\uparrow$} & \textit{FID$\downarrow$}
 & \textbf{Nut Asm.$\uparrow$} & \textit{FID$\downarrow$} \\
\cmidrule(lr){1-3} \cmidrule(lr){4-5} \cmidrule(lr){6-7} \cmidrule(lr){8-9} \cmidrule(lr){10-11} \cmidrule(lr){12-13}
None (Fixed Camera Only) & 10.67 & \textit{54.46} & 18.00  & \textit{59.45} & 14.00 & \textit{43.83} & 47.33 & \textit{\textbf{25.86}} & 30.67 & \textit{86.28} & 13.33 & \textit{45.13} \\
Depth est.+Repoj. & 2.67 & \textit{80.47} & 20.67 & \textit{61.44} & 9.33 & \textit{68.77} & 42.67 & \textit{61.29} & 31.33 & \textit{93.92} & 10.00 & \textit{61.23} \\
VISTA & 28.00 & \textit{\textbf{48.19}} & 56.00 & \textit{44.95} & 40.67  & \textit{43.99} & 66.67 & \textit{38.60} & \textbf{45.33} & \textit{\textbf{64.57}} & 28.67 & \textit{\textbf{25.96}} \\
\midrule
Untranslated (Domain B) & 1.33 & \textit{74.56} & 0.00  & \textit{58.50} & 0.00 & \textit{48.61} & 2.67  & \textit{76.5} & 44.67 & \textit{80.27} & 0.00 & \textit{99.59} \\
MANGO (Domain B $\rightarrow$ A) & \textbf{30.00} & \textit{50.03} & \textbf{86.00} & \textit{\textbf{32.11}} & \textbf{64.67} & \textit{\textbf{42.45}} & \textbf{71.33} & \textit{37.47} & 13.33 & \textit{100.12} & \textbf{45.33} & \textit{53.88} \\
\midrule
Simulator (Oracle) & 57.33 & -- & 100.00 & -- & 80.67 & -- & 86.67 & -- & 86.00 & -- & 64.00 & -- \\

\bottomrule
\end{tabular*}
\end{table*}

\begin{figure*}[h]
    \centering
    \includegraphics[width=1.9\columnwidth]{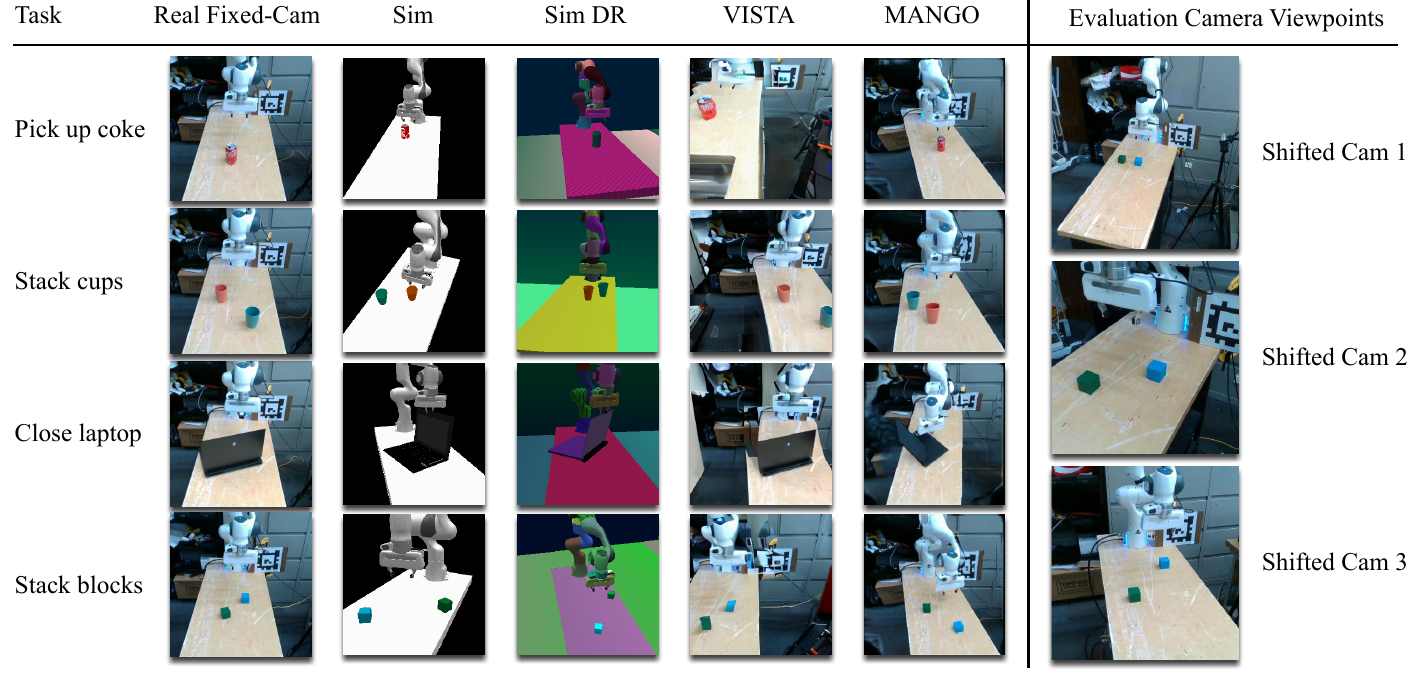}
    \caption{\textbf{Left: }Sample image observations for each task and data-augmentation method from Table \ref{tab:real_results}. \textbf{Right: } The three shifted viewpoints that comprise our ``Shifted Cams'' evaluations in Table \ref{tab:real_results}}
    \label{fig:example_obs}
\end{figure*}

Table \ref{tab:benchmark_results} gives the scores of our proposed method against baselines and ablations. MANGO obtains the lowest FID score by 23 points on the Randomized Camera testset. The patch discriminator $D$ has the largest impact on the FID score for all testsets. Translated image examples are given in Figure \ref{fig:example_obs}.

\begin{table}[H]
    \centering
    \caption{\textbf{Average pairwise LPIPS on natural image datasets.} A core challenge in robot learning is lack of dataset diversity as compared to web data. \\$\dag$ Results computed on $10^3$ randomly sampled images.}
    \label{tab:data_diversity}
    \begin{tabular}{lcp{4cm}}
        \toprule
        Dataset & Average Pariwise LPIPS ($\uparrow$)  \\
        \midrule
        Laion-5B$^\dag$   & 0.725  \\
        ImageNet$^\dag$   & \textbf{0.819}  \\
        Cifar-10$^\dag$  & 0.221    \\
        Cifar-100$^\dag$  & 0.250   \\        
        Horse2zebra $\mathcal{D}_A$ (Horses) &  0.747 \\
        Horse2zebra $\mathcal{D}_B$ (Zebras) &  0.765  \\
        Seg2Cityscapes $\mathcal{D}_B$ (Real)  & 0.548 \\
        \texttt{pick up coke} $\mathcal{D}_B$ (Real) & 0.155 \\

        \bottomrule
    \end{tabular}
\end{table}
\looseness=-1 Note that while the relative FID scores correlate well with relative image quality, the scores in Table \ref{tab:benchmark_results} are high compared to the numbers reported in other literature. We posit that this is due to the small size of our test sets and that our robotics lab scene may be out-of-distribution for the Inception network used for FID Score.

We hypothesize that off-the-shelf methods like CUT and CycleGAN struggle with robotics datasets due to their lack of diversity. Typically, unpaired image-to-image translation methods are benchmarked on computer vision datasets containing diverse images scraped from the internet. In comparison, robotics datasets may be collected from a single setup. To support this claim, we compute the average pairwise LPIPS (a learned perceptual distance metric) \cite{zhang2018unreasonable} on various image datasets as a measure of diversity. As shown in Table \ref{tab:data_diversity}, our $\mathcal{D}_B$ shows the lowest score.

\subsection{Simulation Experiments}

\begin{figure}
    \centering
    \includegraphics[width=0.9\linewidth]{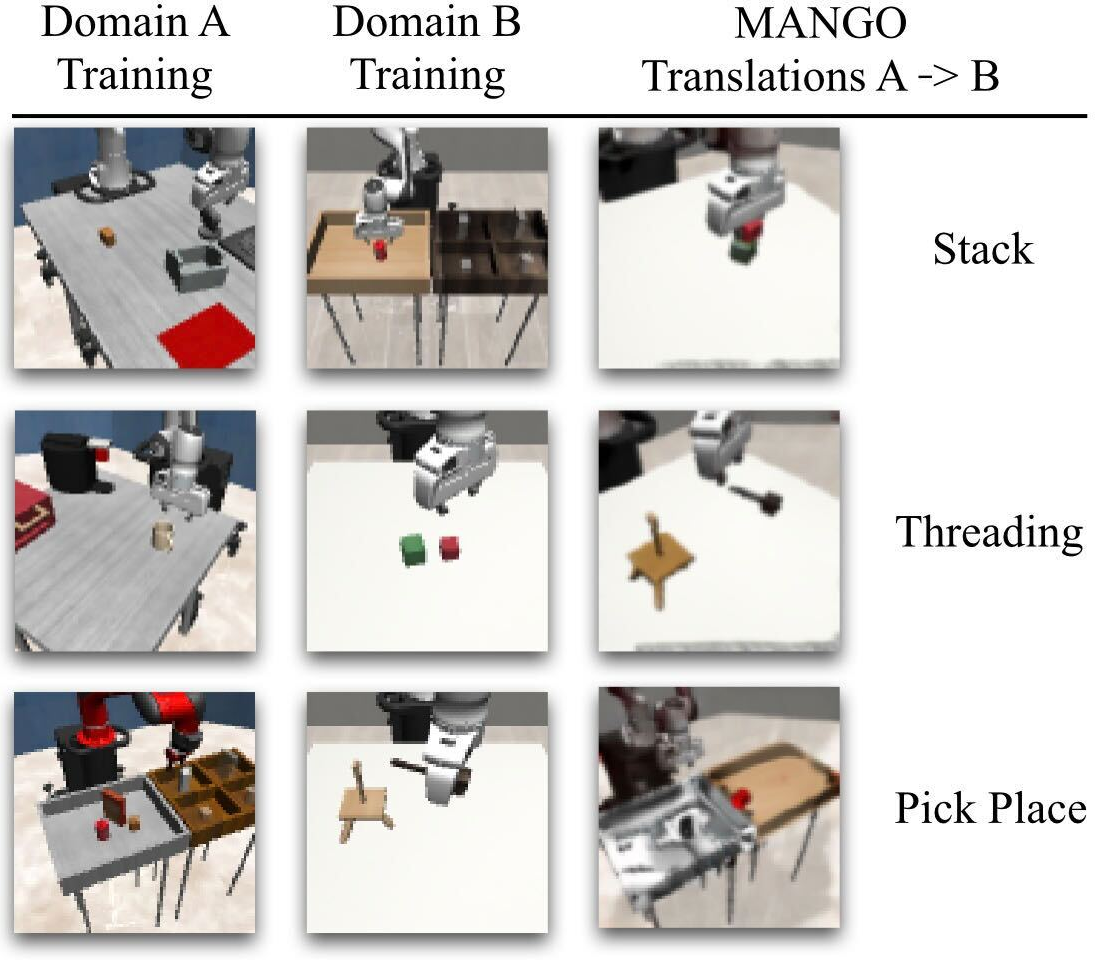}
    \caption{Sim2Sim training data with translations by MANGO for three of the tasks included in Table \ref{tab:sim_results}}
    \label{fig:sim_images}
\end{figure}

We evaluate MANGO on simulated tasks from Robomimic \cite{robomimic2021} and Mimicgen\cite{mimicgen} and give the results in Table \ref{tab:sim_results}.  Specifically, we measure the FID scores of data generated with MANGO, as well as the success rates of behavioral-cloning policies trained with the data. Instead of sim2real, we create two visually disparate simulation environments and run sim2sim experiments. Example observations from each domain and MANGO translations are given in Figure \ref{fig:sim_images}. We benchmark MANGO against policies trained on single-camera observations, untranslated Domain A data, and VISTA \cite{tian2024view}. To provide a fair comparison, we only train MANGO on image observations from the same camera viewpoints and tasks used by VISTA, and report metrics on the same six evaluation tasks. ``Unseen object'' tasks contain objects not seen in training, ``shared object'' contains objects seen during training but in different contexts, and ``cross-embodiment'' tasks are seen performed by the Rethink Sawyer robot instead of Franka Panda. We train MANGO with fixed camera data from the six tasks used for testing in domain B, and varied camera data of the eight training tasks in domain A. This ensures our model does not see views from the test viewpoint distribution, aside from the single fixed view. MANGO-trained BC policies obtain the highest success rate on 5 out of 6 tasks.

\subsection{Real Robot Experiments}
\begin{table*}[hb!]
\centering
\caption{\textbf{Success rates for imitation learning policies across tasks, viewpoint shifts, and data augmentation methods.} The only method comparable to MANGO is VISTA, which uses a 4.5B parameter pretrained model in contrast MANGO's 35M parameters.}
\begin{tabular}{l *{8}{c}}
\toprule
\multirow{2}{*}{\makecell{\textbf{Data}\\\textbf{Augmentation}}}

  & \multicolumn{2}{c}{\textbf{Pick up coke}} 
  & \multicolumn{2}{c}{\textbf{Stack cups}}
  & \multicolumn{2}{c}{\textbf{Close laptop}}
  & \multicolumn{2}{c}{\textbf{Stack blocks}} \\
\cmidrule(lr){2-3}\cmidrule(lr){4-5}\cmidrule(lr){6-7}\cmidrule(lr){8-9}
& \textbf{Fixed Cam} & \textbf{Shifted Cams}
& \textbf{Fixed Cam} & \textbf{Shifted Cams}
& \textbf{Fixed Cam} & \textbf{Shifted Cams}
& \textbf{Fixed Cam} & \textbf{Shifted Cams} \\
\midrule
None & \textbf{8/10} & 5/30 & 8/10 & 1/30 & 10/10 & 19/30 & 8/10 & 1/30 \\
Sim & 6/10 & 14/30 & \textbf{9/10} & 5/30 & 10/10 & 20/30 & \textbf{9/10} & 9/30 \\
Sim DR & \textbf{8/10} & 19/30 & 8/10 & 5/30 & 10/10 & 25/30 & 7/10 & 3/30 \\
VISTA & 7/10 & \textbf{23/30} & 7/10 & 8/30 & 10/10 & \textbf{29/30} & 8/10 & \textbf{18/30} \\
Ours & 6/10 & 17/30 & 8/10 & \textbf{13/30} & 10/10 & 22/30 &\textbf{ 9/10} & 11/30 \\
\bottomrule
\end{tabular}
\label{tab:real_results}
\end{table*}
We benchmark MANGO on four real-world robotic manipulation tasks: ``\texttt{pick up coke}'', ``\texttt{stack cups}'', ``\texttt{close laptop}'', and ``\texttt{stack blocks}''. We train a single image translation model for all tasks, where $\mathcal{D}_A$ and $\mathcal{D}_B$  consist of 59,520 simulated observations and 35,294 fixed-camera real observations from tasks. Our real robot setup is shown in Figure \ref{fig:setup}.

\subsubsection{Policy Training and Evaluation Details}

We train action chunking transformer (ACT) \cite{act} policies on synthetic data generated by MANGO. The generated data is translated from our digital twin demonstrations which leverage the task and motion planner from RLBench \cite{james2019rlbench}. All policies are cotrained with 150 human-teleoperated demos with fixed-camera and wrist camera observations. ACT was chosen to isolate the effects of our generated image data as it does not incorporate any pretraining or language conditioning. We train each ACT policy for 10k epochs with a chunk size of 20. Rollouts are done without temporal aggregation.

We compare ACT models trained on MANGO data to strong sim2real and viewpoint augmentation baselines, depicted in Figure \ref{fig:example_obs}. For domain randomization we follow \cite{pinto2017asymmetric} and randomize color, texture, and lighting for all objects in the scene. VISTA is a viewpoint augmentation method that leverages a fine-tuned ZeroNVS model \cite{tian2024view}. Unlike MANGO, VISTA has built-in rejection sampling for generated images based on LPIPS distance to the original images.

We evaluate each trained policy on 10 variations per task on four different camera viewpoints. The first camera viewpoint is the fixed-camera which the real demonstrations and MANGO training data are collected from. The three shifted viewpoints, depicted in Figure \ref{fig:example_obs}, are aggregated into the ``Shifted Cam'' column in Table \ref{tab:real_results}.

\subsubsection{Results}
Results for MANGO-trained policies and baselines are given in Table \ref{tab:real_results}. With MANGO, we observe large increases in for viewpoint robustness as compared to models trained on fixed-camera human demonstrations only. We also observe that our method is necessary to bridge the sim2real gap, since the policies trained on sim demos without translation perform consistently worse than MANGO.

\section{Discussion and Conclusion}

We propose a novel image generation method, MANGO, which is trained on simulated and real robot data. We observe that with only fixed-camera real data, our novel SegNCE loss, discriminator design, and modified PatchNCE loss enable translation of unseen-viewpoint simulated observations to realistic real-world observations. MANGO-generated data improves the robustness of downstream imitation learning policies to camera shift, as demonstrated by greatly increased success rates six simulated and four real-world manipulation tasks. We observe that our method is superior for sim2real translation in this setting, beating all other image translation methods we tested. 

There are several limitations to this work. MANGO still requires a small amount of real-world data from the evaluation domain for training. Additionally, we struggle to beat VISTA in 3 out of 4 real world tasks when evaluated on the shifted camera viewpoints. The primary benefit of MANGO is its ability to preserve image quality during translations with a lightweight model that is practical for translating robot demonstration datasets with hundreds of thousands of image observations. MANGO, including data generation and image translation, requires less than 0.2\% of the GPU-hours required by VISTA. However, larger pretrained models  inherit a more general understanding of 3D geometry and scenes. Incorporating MANGO's novel loss formulations, specifically the segmentation-based InfoNCE loss, into heavier pretrained models for sim2real visual observation translation or real2real augmentation is a promising direction for future work.

\bibliographystyle{IEEEtran}
\bibliography{references}

\end{document}